%% file: arxiv.tex
\pdfoutput=1

\documentclass[11pt]{article}

\usepackage[preprint]{acl}

\usepackage{times}
\usepackage{latexsym}
\usepackage{booktabs}
\usepackage{hyperref}
\usepackage{makecell}
\usepackage{color}
\usepackage{fancybox}
\usepackage{pifont}
\newcommand{\cmark}{\ding{51}}
\newcommand{\xmark}{\ding{55}}

\usepackage[T1]{fontenc}

\usepackage[utf8]{inputenc}

\usepackage{microtype}

\usepackage{inconsolata}

\usepackage{graphicx}

\usepackage{threeparttable}
\usepackage{tabularx}
\usepackage{xspace}
\usepackage{mdframed}
\usepackage{minted}
\usepackage{multirow}
\usepackage{array}
\usepackage{marvosym}
\usepackage{xcolor}
\usepackage{soul}
\usepackage[most]{tcolorbox}
\usepackage{subcaption}
\usepackage{todonotes}
\usepackage{enumitem}

\usepackage{eso-pic}
\usepackage{graphicx}
\usepackage{geometry}

\definecolor{lightred}{RGB}{255, 200, 200}
\definecolor{lightgreen}{RGB}{200, 255, 200}
\definecolor{deepviolet}{RGB}{96, 48, 96}

\tcbset{
  promptbox/.style={
    top=10pt,
    colback=white,
    colframe=violet,
    colbacktitle=deepviolet,
    fontupper=\itshape,
    enhanced,
    center,
    attach boxed title to top left={yshift=-0.1in,xshift=0.15in},
    boxed title style={boxrule=0pt,colframe=white},
  }
}
\newtcolorbox{promptframed}[2][]{promptbox,title=#2,#1}

\tcbset{
  promptbox1/.style={
    colback=white,
    colframe=violet,
    fontupper=\itshape,
    enhanced,
    center,
  }
}
\newtcolorbox{promptframed1}[2][]{promptbox1,title=#2,#1}

\newcounter{promptcounter}
\newenvironment{prompt}[1]
{
  \refstepcounter{promptcounter}\par\medskip
  \begin{promptframed}{Prompt~\thepromptcounter: #1}
}
{
  \end{promptframed}
}

\AddToShipoutPictureBG*{%
  \AtPageUpperLeft{%
    \raisebox{-1.5cm}[0pt][0pt]{%
      \makebox[\paperwidth]{%
        \begin{minipage}{0.9\paperwidth}
          \centering
          \color{red}%
          \normalsize
          This paper has been accepted for EMNLP 2025 Main Conference. Please cite the official version: \url{https://aclanthology.org/2025.emnlp-main.322/}\\
        \end{minipage}
      }%
    }%
  }%
}

\title{Legal Fact Prediction: The Missing Piece in Legal Judgment Prediction}

\newcommand{\hit}{$^{1}$}
\newcommand{\osaka}{$^{2}$}
\newcommand{\zju}{$^{3}$}
\newcommand{\tsinghua}{$^{4}$}

\author{
  {\hit Junkai Liu\thanks{These authors contributed equally to this work and share first authorship.}, \zju Yujie Tong\footnotemark[1], \hit Hui Huang, \hit Bowen Zheng, \tsinghua Yiran Hu,}\\ 
  \textbf{\zju Peicheng Wu, \osaka Chuan Xiao, \osaka Makoto Onizuka, \hit Muyun Yang\thanks{These authors share corresponding authorship: Muyun Yang <\href{mailto:yangmuyun@hit.edu.cn}{yangmuyun@hit.edu.cn}>, Shuyuan Zheng <\href{mailto:zheng@ist.osaka-u.ac.jp}{zheng@ist.osaka-u.ac.jp}>.}, \osaka Shuyuan Zheng\footnotemark[2]} \\ 
  \hit Harbin Institute of Technology, \osaka The University of Osaka,\\ 
  \zju Zhejiang University, \tsinghua  Tsinghua University
}

\begin{document}

\maketitle

\def\thefootnote{}\footnotetext{Our code and data are available at \url{https://github.com/teijyogen/LFP}.}
\def\thefootnote{\arabic{footnote}}

\input{content/abstract}

\input{content/intro}

\input{content/task}

\input{content/dataset}

\input{content/experiments}

\input{content/related}

\input{content/conclusion}

\input{content/limitation}

\input{content/ethical}

\input{content/acknowledge}

\bibliography{custom}

\input{content/appendix}

\end{document}

%% file: content/abstract.tex
\begin{figure*}[t]
    \centering
    \includegraphics[width=0.86\linewidth]{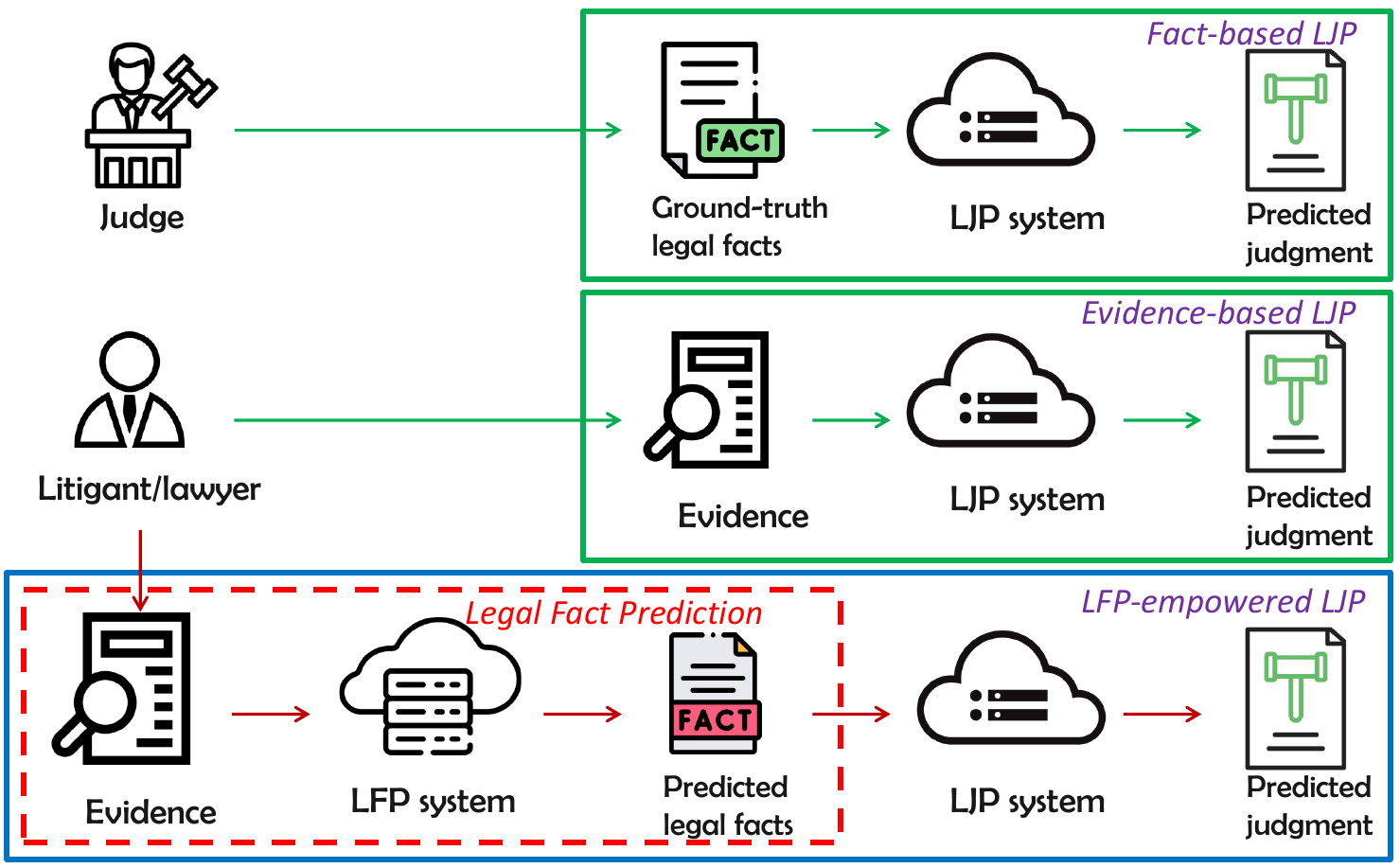}
    \caption{Connection between the legal fact prediction (LFP) and legal judgment prediction (LJP) tasks and comparison of three pipelines of LJP: \textit{fact-based LJP}, \textit{evidence-based LJP}, and \textit{LFP-empowered LJP}. Most existing studies focus on fact-based LJP, while evidence-based LJP and LFP-empowered LJP remain unexplored.}
    \label{fig:overview}
\end{figure*}

\begin{abstract}
Legal judgment prediction (LJP), which enables litigants and their lawyers to forecast judgment outcomes and refine litigation strategies, has emerged as a crucial legal NLP task.
Existing studies typically utilize legal facts, i.e., facts that have been established by evidence and determined by the judge, to predict the judgment. 
However, legal facts are often difficult to obtain in the early stages of litigation, significantly limiting the practical applicability of fact-based LJP. 
To address this limitation, we propose a novel legal NLP task: \textit{legal fact prediction} (LFP), which takes the evidence submitted by litigants for trial as input to predict legal facts, thereby empowering fact-based LJP technologies to make predictions in the absence of ground-truth legal facts.
We also propose the first benchmark dataset, LFPBench, for evaluating the LFP task.
Our extensive experiments on LFPBench demonstrate the effectiveness of LFP-empowered LJP and highlight promising research directions for LFP.
\end{abstract}

%% file: content/intro.tex
\section{Introduction}
\label{sec:intro}

Advancements in NLP technology have significantly propelled the development of legal technology, particularly in the field of legal judgment prediction (LJP). 
LJP aims to predict court rulings based on litigation case information and legal provisions. 
For judges, automated predictions can serve as a reference for their official rulings, ensuring consistency in judicial standards. 
For litigants and their lawyers, pre- or in-trial judgment predictions help assess the potential outcomes of litigation, enabling them to make informed decisions.
Consequently, LJP holds great potential for enhancing judicial efficiency and transparency.

Extensive research efforts have been devoted to achieving accurate LJP.
However, existing LJP research is mostly limited to \textit{fact-based LJP}~\citep{luo2017learning, zhong2018legal, chen2019charge, yue2021neurjudge, feng2022legal, wu2022towards, gan2023exploiting}, where the input to the LJP system consists of (ground-truth) \textit{legal facts}, i.e., facts that are formally established through evidence and determined by the judge.
However, the users of LJP, such as litigants and lawyers, typically confirm their legal facts at a very late stage of the litigation~\cite{medvedeva2023legal}.
Consequently, the application of LJP is largely confined, as the users often seek to predict judgments before litigation or in its early stages to develop and adjust litigation strategies or related plans.

To address the limitations of prior LJP studies, this paper proposes a novel legal NLP task: \textit{legal fact prediction} (LFP), which aims to take the evidence submitted by litigants for trial as input and automatically determine relevant legal facts. Building on this foundation, we further introduce \textit{LFP-empowered LJP}, as illustrated in Figure~\ref{fig:overview}. In this approach, users first input available evidence into the LFP system to generate predicted legal facts, which are then used as the basis for the subsequent LJP task. This approach aligns more closely with real-world legal practice.

To further facilitate research on \textit{LFP} and \textit{LFP-empowered LJP}, this paper introduces the first benchmark dataset for LFP, \textit{LFPBench}, which contains evidence items, legal facts, and judgment outcomes collected from 657 litigation cases in China, covering 10 representative types of civil cases. As such, it can be used to evaluate both the LFP and LJP tasks.

We conducted extensive experiments based on \textit{LFPBench}, leveraging both general-domain and legal-domain models. The results reveal that, compared to fact-based LJP, evidence-based LJP, where judgments are predicted solely based on evidence, exhibits a significant drop in accuracy. This suggests that the absence of legal facts has a profound impact on LJP. 
Moreover, LFP-empowered LJP reduces the accuracy drop of evidence-based LJP by 38.5\% on average.
Therefore, we argue that LFP is a crucial missing piece in the LJP task.

We summarize our contributions as follows:
\begin{itemize}[leftmargin=*]
    \item First, we propose a novel task, \textit{legal fact prediction} (LFP), which empowers LJP applications to operate in a wider range of real-world scenarios.
    \item Second, we introduce \textit{LFPBench}, the first benchmark dataset for studying LFP and LFP-empowered LJP, to support related research.
    \item Third, we conduct extensive experiments on LFPBench.
    Our results confirm the critical role of LFP in the LJP task and reveal the limitations of state-of-the-art (SOTA) models in addressing LFP. 
    These findings offer valuable insights and guidance for future research.
\end{itemize}



%% file: content/task.tex
\begin{figure*}[ht]
\centering
\includegraphics[width=0.86\linewidth]{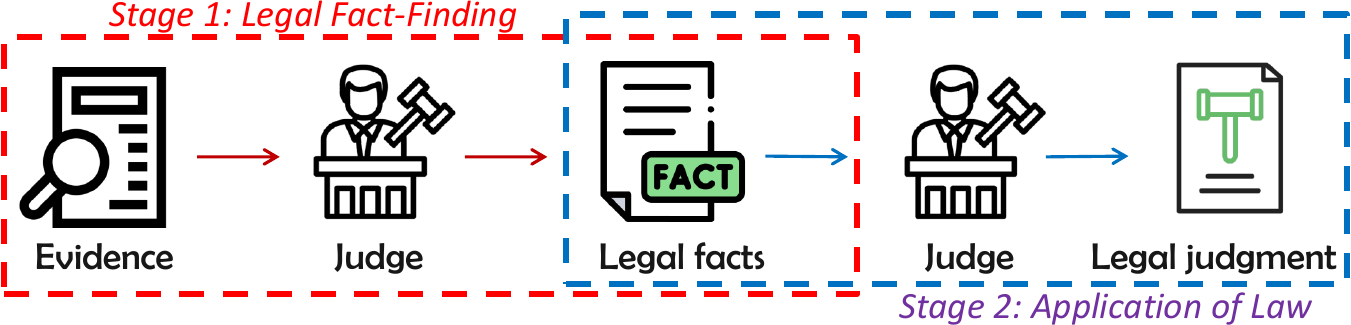}
\caption{A trial primarily addresses two tasks: determining legal facts and applying the law.}
\label{fig:trial}
\end{figure*}

\begin{table*}[t]
\centering
\caption{Comparison between LFPBench and existing LJP benchmarks.}
\resizebox{\textwidth}{!}
{
\begin{tabular}{l|cccccccc}
\hline
\textbf{Benchmark}                                                     & \textbf{\begin{tabular}[c]{@{}c@{}}Evidence\\ items\end{tabular}} & \textbf{Claims}       & \textbf{\begin{tabular}[c]{@{}c@{}}Legal\\ facts\end{tabular}} & \textbf{\begin{tabular}[c]{@{}c@{}}Textual\\ judgments\end{tabular}} & \textbf{\begin{tabular}[c]{@{}c@{}}Judgment\\ labels\end{tabular}} & \textbf{\begin{tabular}[c]{@{}c@{}}Label\\ classes\end{tabular}} & \textbf{Type of cases} & \textbf{Jurisdiction} \\ \hline
\textbf{SwissJP~\cite{niklaus2021swiss}} & \xmark                                                            & \cmark          & \cmark                                                         & \cmark             & \cmark                                                             & 2                                                                                    & Generic                                    & Switzerland                               \\
\textbf{LJP-MSJudge~\cite{ma2021legal}}  & \xmark                                                            & \cmark          & \cmark                                                         & \xmark             & \cmark                                                             & 3                                                                                    & Civil                                      & Mainland China                            \\
\textbf{CAIL2018~\cite{xiao2018cail2018}} & \xmark & \cmark & \cmark & \cmark & \cmark & 2 & Criminal & Mainland China \\
\textbf{ILDC~\cite{malik2021ildc}} & \xmark & \cmark & \cmark & \cmark & \cmark & 2 & Generic & India \\
\textbf{Auto-Judge~\cite{long2019automatic}} & \xmark & \cmark & \cmark & \xmark & \cmark & 2 & Civil & Mainland China \\
\textbf{BrCase~\cite{bertalan2020predicting}} & \xmark & \cmark & \cmark & \xmark & \cmark & 2 & Generic & Brazil \\
\textbf{PhilCases~\cite{virtucio2018predicting}} & \xmark & \cmark & \cmark & \xmark & \cmark & 2 & Criminal & Philippines \\
\textbf{LFPBench (ours)}                                              & \cmark                                                            & \cmark          & \cmark                                                         & \cmark             & \cmark                                                             & 3                                                                                    & Civil                                      & Mainland China                            \\
\hline
\end{tabular}
}
\label{tab:lfp_compare}
\end{table*}

\section{The Legal Fact Prediction Task}
In this section, we provide background information and formally define the LFP task.

\subsection{Background}
In the legal context, \textit{evidence} refers to any material or information used to make the existence of a fact more or less probable~\cite{LII:Evidence}, whereas \textit{legal facts}, also known as findings of fact, are the facts of a case determined by the judge during litigation, based on the presentation and cross-examination of evidence by the parties in a trial~\cite{LII:FindingOfFact}. 
In other words, only facts that can be substantiated by evidence in a court of law can be acknowledged by the judge as legal facts. 

As depicted in Figure \ref{fig:trial}, in civil law countries such as Germany, France, and China, as well as in common law countries like the UK and the US, a trial primarily resolves the following two tasks to reach a judgment: 

\begin{itemize}[leftmargin=*]
    \item \textit{Legal fact-finding}: Given the evidence presented and the arguments made by both the plaintiff and the defendant, the judge determines the legal facts of the case.
    \item \textit{Application of law}: The judge applies the law to the legal facts to assess the validity of the plaintiff's claims and make an appropriate judgment.
\end{itemize}

As evident, legal facts serve as the foundation for the application of law, and before legal fact-finding is complete, it is logically impossible to predict a judgment based on legal facts. 
In fact, legal facts are usually finalized when the judge reaches a judgment.
Therefore, as \citet{medvedeva2023legal} has pointed out, utilizing ground-truth legal facts for LJP is impractical. 
Instead, litigants typically complete evidence collection before litigation or in its early stages, making \textit{evidence-based LJP} a more reasonable choice. However, existing research on LJP primarily assume the accessibility of legal facts, namely they are all limited to \textit{fact-based LJP} (e.g.,~\citep{luo2017learning, zhong2018legal, chen2019charge, yue2021neurjudge, feng2022legal, wu2022towards}), which is mismatched with real-world legal practice.


\subsection{Task Definition}
Next, we introduce the formal definition of the LFP task.
Let $C$ denote the plaintiff's claims, which determine the scope of the trial and constrain the space of legal facts to be predicted. Let $Z$ denote the list of evidence for trial, which records all available evidence items to be presented and examined in establishing legal facts. Then, the LFP task requires a system $f$ that takes the tuple $(C, Z)$ as input and yields a set of legal facts $f(C, Z)$.


The predicted legal facts $f(C, Z)$, along with the claims $C$, can be further input into a given LJP system $g$ to obtain the judgments $g(C, f(C, Z))$ for the claims.
Since our motivation is to enhance evidence-based LJP, the objective of the LFP task is to find the optimal LFP system $f$ that maximizes the accuracy of the predicted judgments $g(C, f(C, Z))$.

Note that the LFP task is not to summarize the evidence into legal facts.
Instead, the available evidence items represent fragmented pieces of legal facts rather than a complete picture. 
Therefore, the LFP task involves deducing and expanding the evidence into legal facts. 
Moreover, conflicts or contradictions may exist among evidence items, particularly between the evidence presented by the plaintiff and the defendant. 
This requires the LFP system to assess the strength and logical coherence of the evidence to resolve the conflicts, making the prediction of legal facts a significant challenge.
Notice the plaintiff's claims and the evidence list are typically available before the trial; therefore, leveraging this information as input for LFP aligns with real-world legal practice.
We discuss the flexible adaptability of the input across different scenarios in Appendix~\ref{sec:discuss}.



%% file: content/dataset.tex
\begin{figure*}[t]
    \centering
    \includegraphics[width=1.0\linewidth]{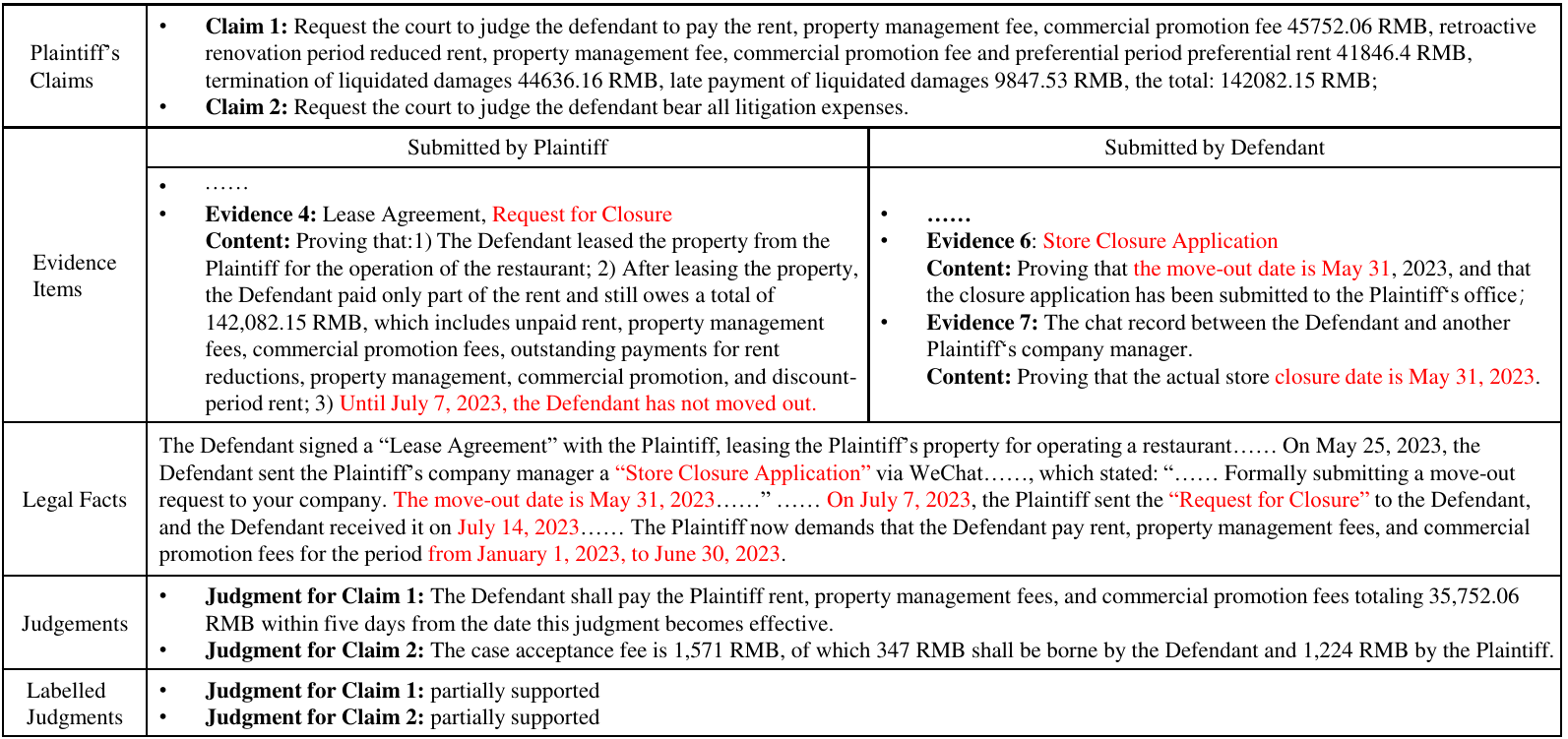}
    \caption{A data sample from the LFPBench dataset featuring a house lease case. Both the plaintiff and the defendant submitted evidence to assist the judge in determining the legal facts. However, Evidence 4, Evidence 6, and Evidence 7 present conflicting information regarding the defendant's actual move-out date. Ultimately, according to Evidence 7, the judge determined that the defendant had not moved out before July 7 and had defaulted on the rent for June.}
    \label{fig:example}
\end{figure*}

\begin{table}[t]
\centering
\small
\caption{Data statistics of the LFPBench dataset. Complete-win: all claims of the plaintiff are supported. Partial-win: part of the plaintiff's claims are completely or partially supported. Loss: all claims of the plaintiff are rejected.}
\resizebox{\linewidth}{!}{
\begin{tabular}{lr|lr}
\toprule
\multicolumn{2}{l|}{No. of cases} & \multicolumn{2}{l}{No. of evid. items (plaintiff)} \\ 
\midrule
Total & 657 & Max & 19 \\
With Defendant Evid. & 387 & Avg. & 4.26 \\
With Third-Party Evid. & 80 & Median & 4 \\ 
\midrule
\multicolumn{2}{l|}{No. of cases results} & \multicolumn{2}{l}{No. of evid. items (defendant)} \\ 
\midrule
Complete-win cases & 166 & Max & 14 \\
Partial-win cases & 397 & Avg. & 1.83 \\
Loss cases & 94 & Median & 1 \\ 
\midrule
\multicolumn{2}{l|}{No. of judgment} & \multicolumn{2}{l}{No. of claims} \\ \midrule
Full support & 631 & Max & 9 \\
Partial support & 621 & Avg. & 2.48 \\
Reject & 378 & Median & 2 \\ 
\bottomrule
\end{tabular}
}
\label{tab:lfpbench_stat}
\end{table}

\begin{figure}[t]
    \centering
    \includegraphics[width=\linewidth]{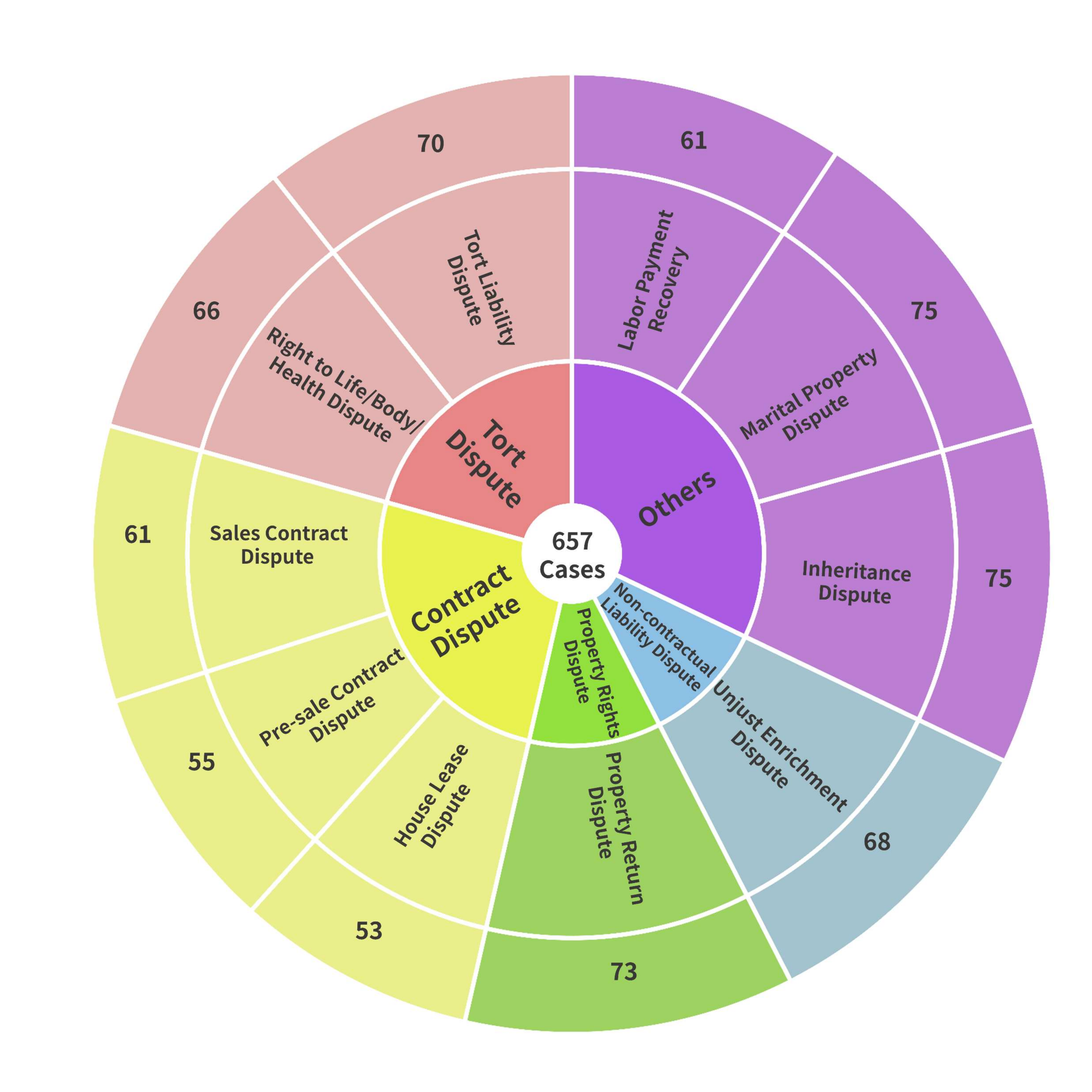}
    \label{fig:piechart}
    \caption{Distribution of case types in the LFPBench dataset.}
    \label{fig:lfpbench_stat}
\end{figure}

\section{LFPBench}

\subsection{Dataset Overview}
As shown in Table~\ref{tab:lfp_compare}, existing LJP datasets, such as CAIL2018~\cite{xiao2018cail2018} and ILDC~\cite{malik2021ildc},  focus on the prediction tasks of prison terms and charges rather than including the evidence items submitted by litigants, making LFP infeasible~\cite{cui2023survey}. 
Therefore, we propose the first benchmark dataset for the LFP task, LFPBench. 
LFPBench consists of data for 657 first-instance cases in China, covering ten types of civil causes of action as shown in Figure~\ref{fig:lfpbench_stat}. 
Each case includes the plaintiff's claims, the evidence items submitted by the litigants, ground-truth legal facts, ground-truth judgments for the claims, and more. 
Therefore, LFPBench can be used for evaluating both LFP and LJP tasks. 
We have selected some widely used datasets for comparison with LFPBench, as shown in the Table~\ref{tab:lfp_compare}. Not only does LFPBench include a third category for partial support by the court (the other two works only include support or opposition), but the input length also far exceeds theirs, posing a higher challenge to the model's capabilities. More importantly, to the best of our knowledge, LFPBench is the only dataset that considers predictions made before the trial in real-world scenarios, thus featuring a unique input of evidence lists, while other benchmark datasets follow the paradigm of using legal facts for judgment prediction.

The data statistics of LFPBench can be found in Table~\ref{tab:lfpbench_stat}. 
In LFPBench, defendants present counterevidence in 58.9\% of cases, leading to disputes over the determination of legal facts. 
Consequently, only 38.71\% of claims are fully supported, and plaintiffs completely win only 25.27\% of the cases. 
Therefore, predicting the legal facts and judgments of these cases is highly challenging.
Figure~\ref{fig:example} presents a data sample from a house lease case, illustrating how conflicting evidence between the litigants complicates the determination of legal facts.

\subsection{Dataset Construction}
LFPBench data was extracted from judicial judgments in China. 
Our legal experts selected ten representative civil litigation causes of action with a moderate level of difficulty in establishing legal facts and retrieved 100 written judgments for each type from the China Judgments Online database~\cite{Wenshu:CourtDocumentsChina}. 
These case types encompass various common disputes over property and personal rights in daily life. 
To ensure quality, three legal experts reviewed the judgments and excluded those with overly vague descriptions of evidence. Ultimately, 657 cases were retained. 

Then, we used regular expressions to extract legal facts and judgment outcomes, as they are typically written in a consistent structure in the documents. 
Conversely, since the writing format for claims and evidence varies across judgments, we employed GPT-4o to extract this information. 
Afterward, our legal experts conducted a manual review to ensure consistency among the extracted claims, evidence, legal facts, and judgments.

Finally, our legal experts annotated the judgment outcomes. 
Specifically, as shown in Figure~\ref{fig:example}, they aligned each claim with its corresponding judgment and categorized the outcome into three labels based on the level of support: fully supported, partially supported, and rejected. 
These labels enable us to evaluate the LFP task using classification-based assessment methods. 

\begin{table}[t]
\centering
\caption{Human evaluation of the quality of the extracted evidence in LFPBench, conducted by two legal experts. For comparison, the combined rates of affirmation and withdrawal in second-instance civil cases in China across different years are presented.}
\resizebox{1.0\linewidth}{!}
{
\begin{tabular}{cl|c}
\hline
\multicolumn{2}{c|}{\textbf{Metric}}                                                                                                      & \textbf{Value} \\ \hline
\multicolumn{2}{c|}{\begin{tabular}[c]{@{}c@{}}Accuracy of Evidence-Based LJP \\ by \textbf{Legal Experts}\end{tabular}} & 87.62\%        \\ \hline
\multicolumn{2}{c|}{Affirmation + Withdrawl Rate in 2022}                                                                                   & 74.78\%        \\
\multicolumn{2}{c|}{Affirmation + Withdrawl Rate in 2023}                                                                                   & 75.61\%        \\
\multicolumn{2}{c|}{Affirmation + Withdrawl Rate in 2024}                                                                                   & 76.63\%        \\ \hline
\end{tabular}
}
\label{tab:h_eval1}
\end{table}

\subsection{Human Evaluation for Extracted Evidence}

In many major jurisdiction‌s such as China, Germany, France, and Japan, access to original evidentiary materials is, in principle, restricted to the parties involved in the case due to privacy considerations. Therefore, we opted to extract evidentiary information from publicly available judicial documents. Although such extracted evidence may lack some details compared to the original materials, the purpose and core content of the evidence are generally faithfully reflected in the judgment texts. 

Additionally, we asked our two legal experts to perform evidence-based LJP using our dataset, i.e., to predict the judgment outcomes solely based on the extracted evidence. They were required to evaluate a randomly selected sample of 100 cases, comprising a total of 259 plaintiff claims. As shown in Table~\ref{tab:h_eval1}, the experts achieved an LJP accuracy of 87.26\%. This level of accuracy is notably high, considering that real-world judicial decisions are not entirely error-free. According to statistics released by the Supreme Court of China, the combined rates of affirmation and withdrawal in second-instance civil cases nationwide were 74.78\%, 75.61\%, and 76.63\% in 2022, 2023, and 2024, respectively~\cite{Gongbao:JudicialStatsChina}.
These findings suggest that our extracted evidence retains the vast majority of critical information, enabling legal experts to make accurate judgments accordingly.

%% file: content/experiments.tex
\begin{table*}[t]
\centering
\small
\caption{Accuracy (\%) of predicted judgments under different LJP approaches and models. \textbf{Def.}: cases where both parties have submitted evidence. \textbf{No def.}: cases where the defendant has not submitted evidence.}
\resizebox{\textwidth}{!}
{
\begin{tabular}{c|ccc|ccc|ccc}
\hline
\multirow{2}{*}{\textbf{Model}} & \multicolumn{3}{c|}{\textbf{Evidence-based}} & \multicolumn{3}{c|}{\textbf{LFP-empowered}} & \multicolumn{3}{c}{\textbf{Fact-based}} \\ \cline{2-10} 
 & All & Def. & No def. & All & Def. & No def. & All & Def. & No def. \\ \hline
GPT-4o & 50.67 & 50.91 & 50.31 & \textbf{51.47} & 49.39 & 54.67 & 55.77 & 52.02 & 61.53 \\
Claude3.5 & 50.80 & 46.36 & 57.63 & \textbf{52.58} & 48.58 & 58.72 & 56.44 & 51.21 & 64.49 \\
Qwen2.5-14B & 45.09 & 42.21 & 49.53 & \textbf{48.10} & 43.83 & 54.67 & 49.45 & 44.74 & 56.70 \\
Llama3.1-Chinese-8B & 40.31 & 34.72 & 48.91 & 40.49 & 34.62 & 49.53 & 40.18 & 35.53 & 47.35 \\
\textbf{Average (General)} & 46.72 & 43.55 & 51.60 & 48.16 & 44.11 & 54.40 & 50.46 & 45.88 & 57.52 \\ \hline
Law-Llama3.1-8B & 31.10 & 28.85 & 35.36 & 30.12 & 26.72 & 40.65 & 33.13 & 32.89 & 33.49 \\
LawJustice-Llama3.1-8B & 35.21 & 30.97 & 41.74 & 28.96 & 26.42 & 32.87 & 32.33 & 26.42 & 41.43 \\
\textbf{Average (Legal)} & 33.16 & 29.91 & 38.55 & 29.54 & 26.57 & 36.76 & 32.73 & 29.66 & 37.46 \\ \hline
\end{tabular}
}
\label{tab:approach_compare}
\end{table*}

\begin{table}[t]
\centering
\caption{LFP similarities under different models.}
\resizebox{1.0\linewidth}{!}
{
\begin{tabular}{l|ccc}
\hline
                             & \textbf{ROUGE} & \textbf{ChatLaw} & \textbf{LLM-as-Judge} \\ \hline
\textbf{GPT-4o}              & 0.1808         & 0.7629           & 5.52                  \\
\textbf{Claude3.5}           & 0.2138         & 0.7668           & 5.83                  \\
\textbf{Qwen2.5-14B}         & 0.1692         & 0.7464           & 5.67                  \\
\textbf{Llama3.1-Chinese-8B} & 0.1763         & 0.7549           & 6.23                  \\
\textbf{LawLlama3.1-8B}      & 0.1785         & 0.7455           & 5.46                  \\
\textbf{LawJustice-8B}       & 0.1721         & 0.7069           & 3.70                  \\ \hline
\end{tabular}
}
\label{tab:lfp_eval}
\end{table}

\section{Experiment}
\label{sec:exp}


\subsection{Setup}

\paragraph{Research questions.}
We conduct experiments to answer the following questions.
\begin{itemize}[leftmargin=*]
    \item RQ1 (Model \& LJP Approach Comparison):  How do SOTA models perform on the LFP and LJP tasks? How do different LJP approaches, including evidence-based LJP, fact-based LJP, and LFP-empowered LJP, perform?
    \item RQ2 (Challenge \& Bias Analysis): What are the challenges of the LFP task? What biases do existing models exhibit when performing LFP?
\end{itemize}

\paragraph{Models.}
We employ 6 LLMs as the LFP and LJP systems, including the closed-source, general-purpose LLMs GPT-4o~\cite{gpt4o} and Claude-3.5-Sonnet-20241022~\cite{Anthropic:ClaudeModels}, the open-source, general-purpose LLMs Qwen2.5-14B-Instruct~\cite{yang2024qwen2} and Llama3.1-Chinese-8B~\cite{llama3.1}, as well as the open-source legal LLMs LawJustice-Llama3.1-8B~\cite{lawjustice} and Law-Llama3.1-8B~\cite{llamalaw}.
Additionally, we evaluated other legal LLMs including DISC-LawLLM~\cite{yue2024lawllm}, Lawyer-Llama-13B-V2~\cite{Lawyer-LLama} and AIE-51-8-Law-Model~\cite{AIE}. However, these models failed to perform the LFP task due to poor instruction-following capabilities, as detailed in Appendix~\ref{app:legalmodels}.

\paragraph{LJP approaches.}
We compare the following LJP approaches that differ in their input, with their prompts detailed in Appendix~\ref{sec:prompts}.
\begin{itemize}[leftmargin=*]
    \item \textit{Evidence-based LJP}: The submitted evidence and the plaintiff's claims are directly input into the LJP system to generate legal judgments.
    \item \textit{LFP-empowered LJP}: The submitted evidence and the plaintiff's claims are first input into the LFP system to predict legal facts. The predicted facts, along with the original inputs, are then fed into the LJP system to generate legal judgments.
    \item \textit{Fact-based LJP}: The LJP system predicts legal judgments based on the ground-truth legal facts and the plaintiff's claims.
\end{itemize}

\begin{figure}[t]
\centering
\includegraphics[width=\linewidth]{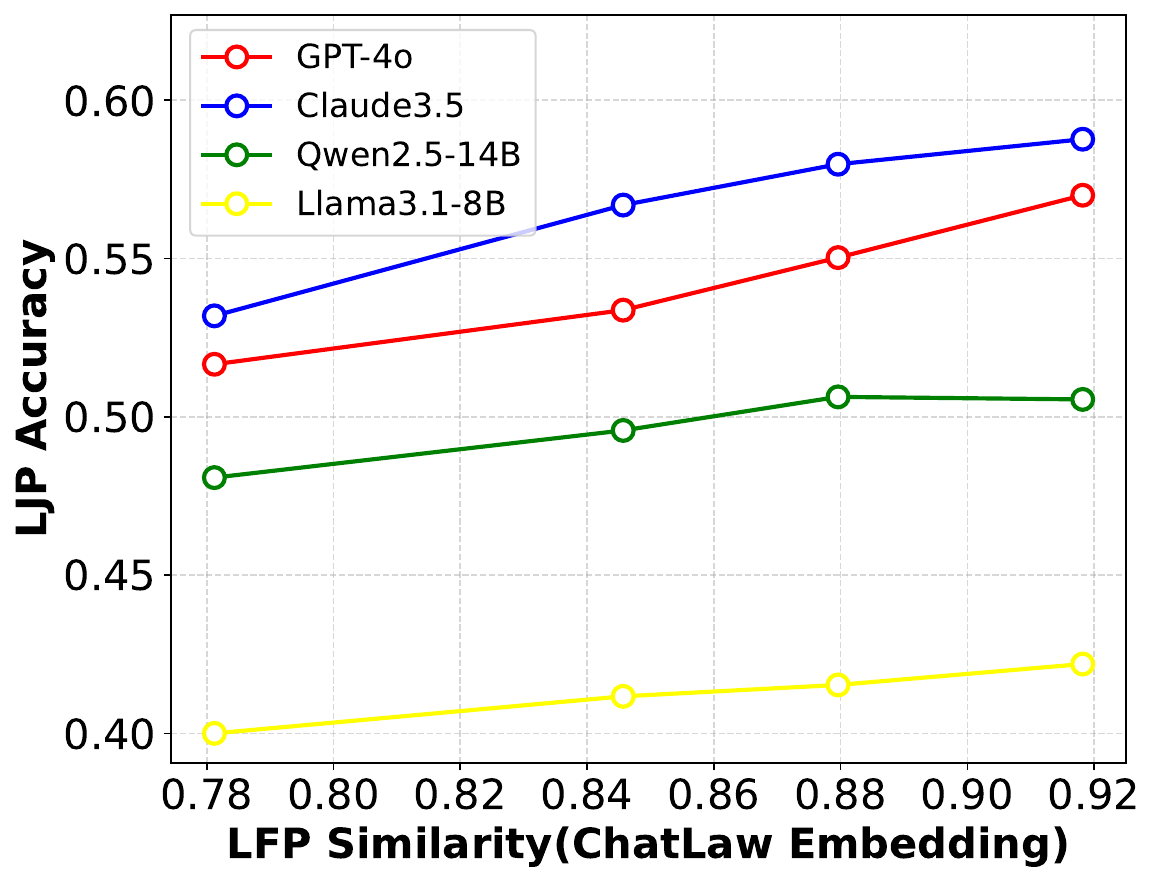}
\caption{The correlation between the LFP similarity and the LJP accuaracy. We leverage the DP-Prompt method~\cite{utpala2023locally} to generate rewritten legal facts with varying LFP similarities.}
\label{fig:noise}
\end{figure}


\begin{table*}[t]
\centering
\caption{Accuracy (\%) of LFP-empowered LJP for different judicial cases. \textbf{Def.}: cases where both parties have submitted evidence. \textbf{No def.}: cases where the defendant has not submitted evidence.}
\small
\resizebox{\textwidth}{!}
{
\begin{tabular}{c|ccc|ccc|ccc}
\hline
\multirow{2}{*}{\textbf{Model}} & \multicolumn{3}{c|}{\textbf{Complete-win case}} & \multicolumn{3}{c|}{\textbf{Partial-win case}} & \multicolumn{3}{c}{\textbf{Loss case}} \\ \cline{2-10} 
 & All & Def. & No def. & All & Def. & No def. & All & Def. & No def. \\ \hline
GPT-4o & 66.95 & 31.06 & 88.29 & 44.49 & 47.85 & 38.02 & 60.47 & 71.52 & 29.82 \\
Claude3.5 & 84.75 & 68.94 & 94.14 & 43.83 & 44.13 & 43.25 & 42.79 & 51.27 & 19.30 \\
Qwen2.5-14B & 75.99 & 60.61 & 85.14 & 45.15 & 45.99 & 43.53 & 16.74 & 20.25 & 7.02 \\
Llama-3.1 & 75.42 & 59.09 & 85.14 & 34.31 & 34.38 & 34.16 & 13.49 & 15.19 & 8.77 \\
Law-Llama3.1-8B & 51.69 & 40.15 & 58.56 & 31.20 & 29.23 & 34.99 & 5.12 & 4.43 & 7.02 \\
LawJustice-Llama3.1-8B & 41.81 & 40.15 & 42.79 & 30.16 & 29.23 & 31.96 & 1.86 & 2.53 & 0.00 \\
\textbf{Average} & \textbf{66.10} & \textbf{50.00} & \textbf{75.68} & \textbf{38.19} & \textbf{38.47} & \textbf{37.65} & \textbf{23.41} & \textbf{27.53} & \textbf{11.99} \\ \hline
\end{tabular}
}
\label{tab:vary_case_types}
\end{table*}

\begin{table*}[t]
\centering
\small
\caption{Accuracy (\%) of LFP-empowered LJP for different causes of action. \textbf{LPR}: Labor Payment Recovery. \textbf{PC}: Pre-sale Contract. \textbf{SC}: Sales Contract. \textbf{ID}: Inheritance. \textbf{HL}: House Lease. \textbf{TL}: Tort Liability. \textbf{UE}: Unjust Enrichment. \textbf{PR}: Property Return. \textbf{MP}: Marital Property. \textbf{RLBH}: Right to Life/Body/Health.}
\resizebox{\textwidth}{!}
{
\begin{tabular}{c|cccccccccc}
\hline
\textbf{Model} & \textbf{LPR} & \textbf{PC} & \textbf{SC} & \textbf{ID} & \textbf{HL} & \textbf{TL} & \textbf{UE} & \textbf{PR} & \textbf{MP} & \textbf{RLBH} \\ \hline
GPT-4o & 60.16 & 52.10 & 62.84 & 50.58 & 50.00 & 54.60 & 53.59 & 53.23 & 39.88 & 40.40 \\
Claude3.5 & 66.41 & 61.08 & 63.51 & 55.81 & 54.49 & 45.98 & 51.63 & 51.08 & 38.15 & 41.72 \\
Qwen2.5-14B & 71.88 & 53.89 & 63.51 & 47.09 & 56.18 & 40.80 & 41.18 & 41.94 & 39.88 & 30.46 \\
Llama3.1-Chinese-8B & 63.28 & 45.51 & 52.03 & 45.35 & 39.33 & 34.48 & 32.03 & 33.87 & 31.21 & 34.44 \\
Law-Llama3.1-8B & 47.66 & 22.75 & 41.22 & 31.98 & 30.34 & 33.91 & 29.41 & 24.73 & 33.53 & 31.79 \\
LawJustice-Llama3.1-8B & 42.19 & 26.95 & 35.81 & 26.74 & 24.72 & 29.31 & 24.18 & 22.58 & 34.10 & 27.15 \\ \hline
\textbf{Average} & \textbf{58.60} & \textbf{56.37} & \textbf{53.15} & \textbf{42.93} & \textbf{42.51} & \textbf{39.85} & \textbf{38.67} & \textbf{37.91} & \textbf{36.13} & \textbf{34.33} \\ \hline
\end{tabular}
}
\label{tab:vary_cause}
\end{table*}

\paragraph{Metrics}
We primarily use \textbf{LJP accuracy} as the metric to quantify the influence of LFP on LJP.
To measure \textbf{LFP similarity}—that is, the similarity between predicted and ground-truth legal facts—we employ three metrics:
(1) ROUGE \cite{lin2004rouge};
(2) ChatLaw-based similarity, which calculates the distance between the predicted and ground-truth facts based on their embeddings determined by the ChatLaw-Text2Vecw model~\cite{cui2023chatlaw}; and
(3) LLM-Judge scores assigned by GPT-4o on a ten-point scale.
We also report F1 scores in Appendix~\ref{sec:f1}, which demonstrate consistency across metrics.

\begin{table*}[t]
\centering
\small
\caption{The proportions of cases in which each model predicts a complete win/partial win/loss for the plaintiff using LFP-empowered LJP. Evidence items are ordered, ensuring all evidence from the plaintiff (or defendant) appears first.}
\resizebox{\textwidth}{!}
{
\begin{tabular}{c|cccc|cccc}
\hline
\multirow{2}{*}{\textbf{Model}} & \multicolumn{4}{c|}{\textbf{Defendant first, plaintiff last}} & \multicolumn{4}{c}{\textbf{Plaintiff first, defendant last}} \\ \cline{2-9} 
 & Accuracy & \begin{tabular}[c]{@{}c@{}}Complete\\ win rate\end{tabular} & Loss rate & \begin{tabular}[c]{@{}c@{}}Partial \\ win rate\end{tabular} & Accuracy & \begin{tabular}[c]{@{}c@{}}Complete\\ win rate\end{tabular} & Loss rate & \begin{tabular}[c]{@{}c@{}}Partial \\ win rate\end{tabular} \\ \hline
GPT-4o & 42.19 & 16.26 & 29.75 & 53.99 & 48.80 & 12.27 & 28.83 & 58.90 \\
Claude3.5 & 39.78 & 42.33 & 1.84 & 55.83 & 49.64 & 37.42 & 18.40 & 44.17 \\
Qwen2.5-14B & 45.55 & 29.14 & 1.23 & 69.63 & 44.35 & 26.69 & 2.15 & 71.17 \\
Llama3.1-Chinese-8B & 35.23 & 19.69 & 2.77 & 77.54 & 34.74 & 22.09 & 2.45 & 75.46 \\
Law-Llama3.1-8B & 28.61 & 4.45 & 0.40 & 95.14 & 27.04 & 4.60 & 1.15 & 94.25 \\
LawJustice-Llama3.1-8B & 25.36 & 3.07 & 0.00 & 96.93 & 28.49 & 2.03 & 0.0 & 97.97 \\
\textbf{Average} & \textbf{36.12} & \textbf{19.16} & \textbf{5.99} & \textbf{74.51} & \textbf{38.84} & \textbf{17.52} & \textbf{8.83} & \textbf{73.65} \\ \hline
Ground truth & 100.00 & 17.18 & 17.18 & 65.64 & 100.00 & 17.18 & 17.18 & 65.64 \\ \hline
\end{tabular}
}
\label{tab:evidence_order}
\end{table*}

\begin{figure*}[ht]
    \includegraphics[width=\textwidth]{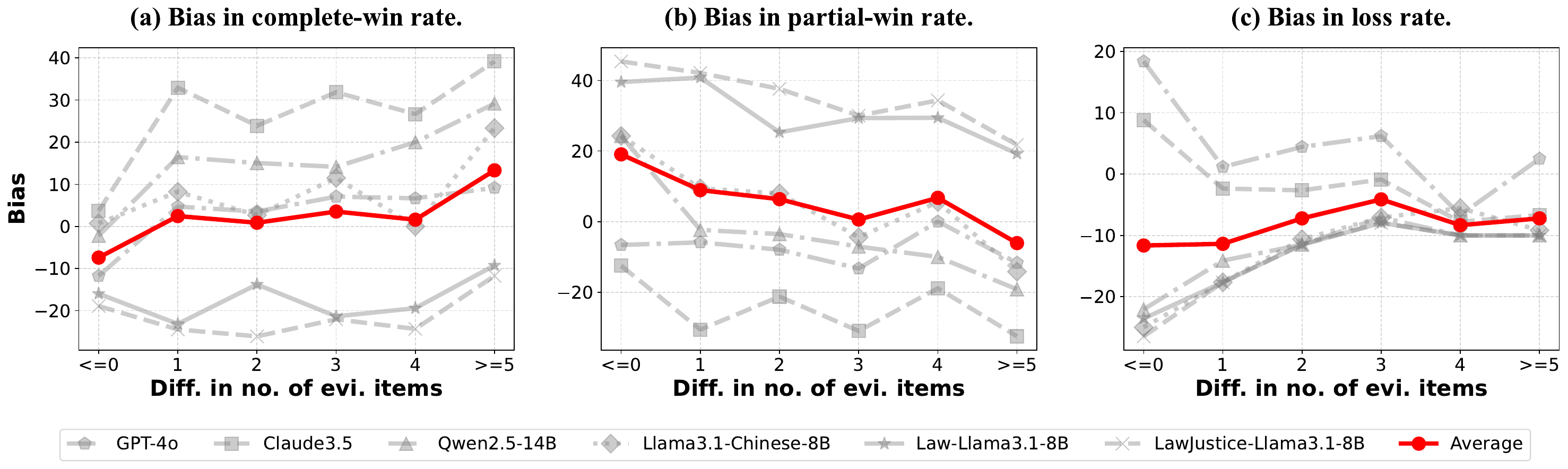}
    \caption{The effect of the difference in the number of evidence items between the plaintiff and the defendant on the judgments yielded by LFP-empowered LJP. The bias in y-axis means the difference in the rate between the model's predictions and the ground truth.}
    \label{fig:bias_in_nb}
\end{figure*}

\subsection{Model \& LJP Approach Comparison (RQ1)}

\paragraph{Finding 1: Legal LLMs perform poorly in LFP and LFP-empowered LJP.}
As shown in Table~\ref{tab:approach_compare}, the closed-source models GPT-4o and Claude3.5 consistently achieve the best performance across different LJP approaches, while the open-source legal LLMs Law-Llama3.1-8B and LawJustice-Llama3.1-8B perform the worst, with accuracy close to random guessing. For the LFP performance in Table~\ref{tab:lfp_eval}, the open-source legal LLMs again perform the worst, with the lowest performance among all metrics.
This discrepancy cannot be attributed solely to the relatively small size of the legal LLMs, as Llama3.1-Chinese-8B, which has the same model size, performs significantly better. 
One possible explanation is that these legal LLMs are typically fine-tuned on short-text legal QA datasets~\cite{lawsft1, yue2024lawllm, Lawyer-LLama}, making them less capable of handling complex tasks such as LFP and LJP, which require summarization, reasoning, and deduction over long texts. 
Future research could explore incorporating general-domain instruction data into fine-tuning to mitigate the catastrophic forgetting of fundamental capabilities. 
It is also promising to develop more complex reasoning datasets in the legal domain. 


\paragraph{Finding 2: Incorporating LFP can substantially reduce the performance gap between evidence-based LJP and fact-based LJP.}
As shown in Table~\ref{tab:approach_compare}, for the general-purpose LLMs, predicting legal judgments directly from evidence results in a \textbf{$\frac{50.46-46.72}{50.46} = 7.42\%$} decrease in accuracy compared to predictions based on ground-truth legal facts.
This indicates that while fact-based LJP research has made considerable progress, its effectiveness heavily relies on the accessibility of legal facts. 
On the other hand, although the more practice-aligned evidence-based LJP underperforms\footnote{Note that for the legal LLMs, evidence-based LJP outperforms the other approaches. However, given their poor performance close to random guessing, this difference is likely due to randomness rather than a meaningful advantage.}, the incorporation of LFP reduces the performance gap by 38.50\%.
Therefore, predicting legal facts from evidence first and then making legal judgments based on the predicted facts can significantly improve the accuracy of evidence-based LJP. 
Compared to fact-based LJP, LFP-empowered LJP accommodates a broader range of LJP scenarios in legal practice, striking a favorable balance between accuracy and applicability.

\paragraph{Finding 3: More accurate legal facts yield more accurate legal judgments.} 
Using the DP-Prompt method~\citet{utpala2023locally}, we rewrite the predicted legal facts with the Qwen2.5-14B-Instruct model, generating four versions with varying levels of LFP similarity.
We then perform LJP on each version of the rewritten facts and repeated the experiment three times for each parameter setting.
As shown in Figure~\ref{fig:noise}, there is a positive correlation between the similarity of the rewritten legal facts and the accuracy of the corresponding legal judgments.
These results further underscore the importance of the LFP task in enhancing LJP performance: more accurate legal facts lead to more accurate legal judgments.


\subsection{Challenge \& Bias Analysis (RQ2)}
\paragraph{Finding 4: Judgement prediction for partial-win or loss cases is more challenging.}
In Table~\ref{tab:vary_case_types}, we report the performance of the LLMs in LFP-empowered LJP across cases where the plaintiff achieves a complete win, a partial win, or a loss. 
The results show that accuracy is typically highest in complete-win cases, while loss cases are the most challenging.
This may be because, in real-world scenarios, complete-win cases usually have strong supporting evidence, making it easier to infer legal facts and judgments.
In contrast, partial-win and loss cases often involve significant disputes between the parties, with conflicting or contradictory evidence, making it difficult to establish legal facts and reach a judgment. 

\paragraph{Finding 5: Judgement prediction for cases with defendant evidence is more challenging.}
In Tables~\ref{tab:approach_compare} and~\ref{tab:vary_case_types}, we distinguish between cases where the defendant did or did not submit evidence.
The results show that LFP-empowered LJP performs better in the former. 
This further suggests that counterevidence presented by the defendant can hinder LLMs' ability to infer legal facts. 
Therefore, future research should focus on enhancing LLMs' reasoning capabilities to better assess the authenticity of information.

\paragraph{Finding 6: Judgment prediction with weaker evidence is more challenging.}
From Table~\ref{tab:vary_cause}, we observe that cases involving LPR, PC, SC, ID, and HL, which typically feature strong written evidence (e.g., contracts and wills), allow for easier prediction of legal facts and judgments. 
Moreover, due to their textual nature, written evidence is more easily understood by LLMs.
In contrast, cases such as TL and RLBH, which involve torts, often rely on non-written evidence, such as physical objects and audiovisual materials. 
When described in text, these forms of evidence lose significant detail. 
This suggests that future research could leverage multimodal technology to better interpret image- and sound-based evidence.

\paragraph{Finding 7: Bias arises from the presentation order of evidence items.}
In Table~\ref{tab:evidence_order}, we select all cases that include both plaintiff and defendant evidence, sort the evidence list in different orders, and then feed them to the models for LFP-empowered LJP. 
We find that the order in which plaintiff and defendant evidence appears significantly influences the predicted judgments, introducing a bias favoring the party whose evidence is presented last. 
Specifically, when plaintiff evidence appears later, the LLMs are more likely to predict a complete or partial win for the plaintiff. Conversely, when defendant evidence appears later, the likelihood of the plaintiff losing the case increases. 
This bias may be attributed to the attention mechanism of LLMs \cite{yu2024mitigatepositionbiaslarge}, which requires further exploration in future research.

\paragraph{Finding 8: Bias arises from the number of evidence items.}
Figure~\ref{fig:bias_in_nb} illustrates the impact of the difference in the number of evidence items between the plaintiff and the defendant on the predicted judgments. 
We observe that as the gap in evidence quantity increases, the LLMs generally tend to predict judgments with a higher complete-win rate and a lower partial-win rate compared to the ground truth. 
This suggests that an advantage in evidence quantity may lead LLMs to develop a bias toward fully supporting the plaintiff. 
However, the effect of this advantage on the loss rate is highly divergent: 
as the plaintiff's advantage increases, the two closed-source models become less likely to predict a loss for the plaintiff, whereas the open-source models exhibit the opposite. 
Nevertheless, more efforts are needed to teach LLMs that more evidence items don't necessarily mean stronger evidence or cause legal facts.

%% file: content/related.tex
\section{Related Work}
\label{sec:related works}

\paragraph{Legal Fact Prediction}
Research on LJP can be traced back to the 1960s~\citep{lawlor1963computers}, which is one of the most fundamental tasks in legal AI. 
As judgment documents have become publicly accessible in many countries, researchers have extracted legal facts and judgment results from these documents, forming plenty of benchmark datasets for LJP research (e.g.,~\citep{xiao2018cail2018, chalkidis2019neural, malik2021ildc, semo2022classactionprediction, chalkidis2022lexglue, hwang2022multi}). 
Using judgment documents, numerous legal NLP studies have explored fact-based LJP, which predicts legal judgments based on legal facts, achieving promising predictive accuracy (e.g.,~\citep{luo2017learning, zhong2018legal, chen2019charge, yue2021neurjudge, feng2022legal, wu2022towards, gan2023exploiting}).
However, legal facts are not objective facts and are often difficult for LJP’s intended users to obtain before a judgment is rendered. 
Consequently, Medvedeva et al.~\citep{medvedeva2023legal} recently pointed out that most existing LJP studies rely on unrealistic input such as legal facts, limiting their practicality.
Several studies employed legal briefs~\citep{tippett2021does}, complaint documents~\citep{mcconnell2021case}, court debate records~\citep{ma2021legal} for LJP, but these types of judicial documents are typically not publicly accessible, making it impractical to obtain large-scale datasets for training LJP models.
To address the lack of practicality in current LJP research, this paper proposes the LFP task as a preliminary step to fact-based LJP. 
LFP-empowered LJP establishes a practical loop from evidence to judgment, thereby making LJP more applicable in real-world scenarios.

\paragraph{Legal Document Summarization}
Legal Document Summarization (LDS) is the most correlated task with our LFP, which aims to automatically producing concise, accurate, and coherent summaries of legal texts~\citep{kanapala2019text}, such as judgment documents~\cite{polsley2016casesummarizer}, contracts~\cite{manor2019plain}, and court debate records~\cite{duan2019legal}.
Intuitively, while LFP enriches and assembles fragmented and concise pieces of evidence into complete legal facts, LDS takes the opposite approach by refining legal texts and eliminating lengthy and complex details.
Additionally, LFP requires inferring logically coherent factual information from conflicting or contradictory evidence, making it more challenging than text summarization.

%% file: content/conclusion.tex
\section{Conclusion}


This paper introduces the LFP task to automate the prediction of legal facts for the subsequent LJP task, addressing recent concerns that using ground-truth legal facts for LJP is impractical.
We constructed a benchmark dataset for the LFP task, LFPBench, based on publicly available judicial documents. 
Extensive experiments conducted on LFPBench reveal that SOTA LLMs struggle to accurately determine legal facts when faced with conflicting or contradictory evidence, and exhibit biases related to the quantity and presentation order of evidence.
Future work includes addressing the above limitations and constructing larger-scale LFP datasets to facilitate more extensive research.

%% file: content/limitation.tex
\section*{Limitations}

As the first step on the LFP task, this work has the following limitations. 
First, the evidence information in LFPBench is extracted from publicly accessible judgment documents in China, which are typically summarized and may lack some details of the original evidence, making it more challenging to predict legal facts. 
Second, the scope of case types covered by LFPBench remains limited, as it does not include criminal or administrative litigation cases.
However, we would like to note that focusing on civil cases within a single jurisdiction is a common practice for benchmarking LJP (see Table~\ref{tab:lfp_compare}).

%% file: content/ethical.tex
\section*{Ethical Considerations}
Our work may raise the following ethical considerations.
(1) Data Privacy and Confidentiality: Judicial documents often contain some basic personal information of litigants, such as names, addresses, and identity numbers. 
Although we have processed the data to remove or anonymize personally identifiable information (PII), we must still comply with data usage regulations and refrain from any de-anonymization attempts that could compromise personal privacy.
(2) Judicial Bias.
Inappropriate applications of LJP may introduce ethical challenges, particularly because current fact-based LJP research often relies on legal facts extracted from court opinions, which may reflect judges or jurors' biases. 
As a result, such biases can be embedded in LJP's decisions. 
However, the introduction of the LFP task offers a way to alleviate this issue: it shifts the predictive foundation of LJP from biased legal facts in court opinions to facts predicted by LFP based on objective evidence. We believe this approach can, to some extent, reduce the influence of judicial bias.
(3) Automated Adjudication.
Some voices have proposed using LJP systems to replace human judges and juries, which has raised ethical concerns. However, we believe that the primary purpose of LFP and LJP is to assist litigants and their lawyers in predicting potential court opinions, thereby enabling them to adjust their strategies accordingly. This application can improve judicial transparency and reduce unnecessary judicial costs.

Additionally, we used ChatGPT to polish the writing and are responsible for all the materials presented in this work.

%% file: content/acknowledge.tex
\section*{Acknowledgements}

This work is supported by the National Natural Science Foundation of China (62276077, 62376075, 62376076), JSPS Kakenhi (JP23K17456, JP23K25157, JP23K28096, JP25H01117, JP25K21207), and
CREST (JPMJCR22M2).

%% file: content/appendix.tex
\clearpage

\appendix
\onecolumn

\input{content/discussion}

\section{Additional Details on LFPBench}
\label{sec:annote}

\subsection{Annotation}
Our dataset's annotators consist of four graduate students, two of whom have academic background in law, and two in computer science. They are all co-authors of this thesis, and therefore, no remuneration was provided. Before beginning the annotation process, the annotators unified the criteria for "supported", "partially supported", and "rejected". Generally speaking, if there is any conflict between the court's ruling and the content of a claim (such as a minor discrepancy in the amount of money), it cannot be considered that the court supports the claim. Correspondingly, if there is any overlap between the court's ruling and the content of a claim (such as the recognition of a small portion of the damages), it cannot be considered that the court rejects the claim.


\subsection{Anonymization}
Before extracting the relevant legal judgment information, we have already removed and replaced all sensitive information, such as the names and identification numbers of the litigants. Therefore, using this benchmark is safe and does not pose any risk of personal information leakage.

\subsection{Copyright Issue}
The case data used in LFPBench is sourced from the China Judgments Online (\url{https://wenshu.court.gov.cn/}), a website established by the Supreme People's Court of China for the publication of judgments issued by courts at all levels in China. According to Article 5 of the Copyright Law of the People's Republic of China: "This Law does not apply to: (1) laws, regulations, resolutions, decisions, orders, and other documents of a legislative, administrative, or judicial nature, and their official translations..." As such, the judgment data used in this paper falls under the exemption outlined in this provision and is not subject to the Copyright Law.

\section{Additional Details on Experiments}

\subsection{Documentation of LLMs}
\begin{table}[h]
\centering
\caption{Documentation of the used LLMs.}
\resizebox{0.9\textwidth}{!}{
\begin{tabular}{lccc}
\toprule
\textbf{Artifacts} & \textbf{License} & \textbf{Parameter Scale} & \textbf{Language}\\
\midrule
Lawyer-Llama-13B-V2 & Apache License 2.0 & 13B & Chinese\\
DISC-LawLLM & Apache License 2.0 & 13B & Chinese\\
Llama3.1-Chinese-8B & Apache License 2.0 & 8B & Chinese\\
LawJustice-Llama3.1-8B & Apache License 2.0 & 8B & Chinese/English\\
Law-Llama3.1-8B & Not specified & 8B & Chinese\\
AIE-51-8-Law-Model & Not specified & 3B & Chinese\\
Qwen2.5-14B & Apache License 2.0 & 14B & Multilingual\\
GPT-4o & Closed-source & Unknown & Multilingual\\
Claude3.5 & Closed-source & Unknown & Multilingual\\
\bottomrule
\end{tabular}
}
\label{tab:license}
\end{table}

Table \ref{tab:license} shows the basic information of the LLMs involved in our paper. The Apache License 2.0 allows us to freely use these assets for academic research. One of the goals in training domain-specific models is to achieve performance comparable to larger models within specific domains using fewer parameters. Therefore, we have opted for a model with a relatively small parameter count, ranging between 8-14B. All these models possess Chinese language capabilities, making them suitable for our benchmark tests.
We report the hyperparameters used for the LLMs in Table~\ref{tab:model_parameters}.


\begin{table}[h]
\centering
\caption{Hyperparameter settings.}
\begin{tabular}{l|cc}  
\toprule
\textbf{parameter} & \textbf{Close-source Models} & \textbf{Open-source Models} \\ \midrule
frequency\_penalty & 0 & 0 \\
logprobs & false & false \\
presence\_penalty & 0 & 0 \\
temperature & 1 & 0.7 \\
max\_output\_tokens & 4,096 & 2048 \\
top\_p & 1 & 0.8 \\ \bottomrule
\end{tabular}
\label{tab:model_parameters}  
\end{table}

\subsection{Selection of Legal LLMs}
\label{app:legalmodels}

We conducted a preliminary evaluation to select legal LLMs. 
Each model was tested on the entire LFPBench dataset to evaluate its performance on the LJP task, and we calculated the percentage of its outputs that could be successfully extracted by the evaluation script. 
If the model's results were recognized by the script, it indicates that the model could follow our instructions for LJP. 

Table~\ref{tab:IFR} reports the performance of the general-purpose LLM Llama3.1-Chinese-8B and five legal LLMs in the preliminary evaluation, including DISC-LawLLM~\cite{yue2024lawllm} and Lawyer-Llama-13B-V2~\cite{Lawyer-LLama}, Law-Llama3.1-8B, LawJustice-Llama3.1-8B, and AIE-51-8-Law-Model~\cite{AIE}.
The results show that while Llama3.1-Chinese-8B could fully follow our instructions, the long and complex instructions posed significant challenges for the legal LLMs.
Finally, we selected the two legal LLMs with the highest success rate in instruction following for the main experiments in Section~\ref{sec:exp}.

\begin{table}[h]
\centering
\caption{Success rate (\%) of different legal domain models in following our LJP instructions. For reference, the general-purpose open-source model Llama3.1-Chinese-8B fully complies with our instructions.}
\begin{tabular}{l|cc}
\toprule
Models & Base Models & Instruction Following Rate (\%) \\ \midrule
Llama3.1-Chinese-8B & Llama3.1-8B (2024) & 100.00 \\
Law-Llama3.1-8B & Llama3.1-8B (2024) & 88.74 \\
LawJustice-Llama3.1-8B & Llama3.1-8B (2024) & 80.06 \\
DISC-LawLLM & Baichuan-13B (2023) & 70.47 \\
Lawyer-Llama-13B-V2 & Llama2-13B (2023) & 62.42 \\
AIE-51-8-Law-Model & Qwen2.5-3B (2024) & 51.09 \\
\bottomrule
\end{tabular}
\label{tab:IFR}
\end{table}

\subsection{F1 Scores}

\label{sec:f1}

We report the macro-average and micro-average F1 scores for various models and methods in Table~\ref{tab:f1}. 
These results align with the accuracy metrics presented in Table~\ref{tab:approach_compare}.

\begin{table}[h]
\centering
\caption{The macro-f1 and micro-f1 metrics of the ternary classification under different LJP approaches and models.}
\begin{tabular}{l|cc|cc|cc}
\hline
\multirow{2}{*}{\textbf{Model}} & \multicolumn{2}{c|}{\textbf{Evidence-based}} & \multicolumn{2}{c|}{\textbf{LFP-empowered}} & \multicolumn{2}{c}{\textbf{Fact-based}} \\ \cline{2-7} 
                                & Macro-F1              & Micro-F1             & Macro-F1             & Micro-F1             & Macro-F1           & Micro-F1           \\ \hline
\textbf{GPT-4o}                 & 0.4957                & 0.5086               & 0.5073               & 0.5150               & 0.5503             & 0.5611             \\
\textbf{Claude3.5}              & 0.4530                & 0.5083               & 0.4914               & 0.5264               & 0.5520             & 0.5780             \\
\textbf{Qwen2.5-14B}            & 0.4109                & 0.4534               & 0.4335               & 0.4840               & 0.4725             & 0.5157             \\
\textbf{Llama3.1-Chinese-8B}    & 0.3356                & 0.4101               & 0.3572               & 0.4125               & 0.3450             & 0.3971             \\
\textbf{Law-Llama3.1-8B}        & 0.2825                & 0.3141               & 0.2859               & 0.3221               & 0.2523             & 0.2822             \\
\textbf{LawJustice-Llama3.1-8B} & 0.2714                & 0.3521               & 0.2289               & 0.2900               & 0.2422             & 0.3092             \\ \hline
\end{tabular}
\label{tab:f1}
\end{table}

\subsection{Prompt Templates}
\label{sec:prompts}
In this section, we present the prompt templates we used for evidence extraction, claim extraction, legal fact prediction, evidence-based LJP and fact-based LJP. We have performed initial adjustment to the prompt templates to ensure the performance of different models.


\begin{prompt}{\small{Evidence Extraction}}
\small
    \textbf{[Court Record]}

    **the original text of the reference judgment paper**

    \textbf{[Evidence List]}

    **the reference evidence list**

    Please follow the format of the example above to extract a list of evidence from the provided trial records and output it in the form of a JSON list. Each element in the list should be a dictionary representing a piece of evidence, containing two key-value pairs: "Party Submitting Evidence" and "Content of Evidence." The Party Submitting Evidence should be one of [Plaintiff, Defendant, Third Party], while the Content of Evidence should be extracted directly from the original text of the trial records.

    \textbf{[Court Record]}

    **the original text of the target judgment paper**

    \textbf{[Evidence List]}
\end{prompt}

\begin{prompt}{\small{Claim Extraction}}
\small
    \textbf{[Court Record]}

    **the original text of the reference judgment paper**

    You need to extract all claims of plaintiff from the court record given above, and then organize them into such a list:

    \textbf{[Plaintiff's Claims]}
    
    [

    "**claim1**",

    "**claim2**",

    ...
    
    ]

    Each claim in the list should be as faithful to the original text as possible.
    Focus on the subjective opinions put forward by the defendant, and do not pay attention to his specific evidence.
    You only need to output the formatted list of claims, without adding any comments.

     \textbf{[Plaintiff's Claims]}
\end{prompt}

\begin{prompt}{\small{Legal Fact Prediction}}
\label{prompt:fact_generation}
\small
    \textbf{[Plaintiff's Claims]}

    (1) **claim1**

    (2) **claim2**

    ...

    \textbf{[Litigant]}

    **the parties concerned**

    \textbf{[Evidence List]}

    (1)  **submitting party**

    **content**

    (2)  **submitting party**

    **content**

    ...
    
    Please analyze the plaintiff's claims and the list of evidence in the above case, and output a faithful description of the basic facts of the case from the court's perspective.
    
    Only provide the findings of fact, without adding any reasoning process or explanations.
\end{prompt}

\begin{prompt}{Evidence-Based LJP}
\small
    \textbf{[Case Type]}

    **one of the ten types**

    \textbf{[Litigant]}

    **the parties concerned**

    \textbf{[Evidence List]}

    (1)  **submitting party**

    **content**

    (2)  **submitting party**

    **content**

    ...

    \textbf{[Plaintiff's Claims]}

    (1) **claim1**

    (2) **claim2**

    ...

    You need to refer to the evidence presented by all parties in the [Evidence List] to predict the court's judgment on the [Plaintiff's Claims], and form a corresponding judgment list.
    
    The [Judgment List] is a list composed of three numbers (0, 1, -1). If you believe the court will fully support the claim, the result is 1; if partially supported, the result is 0; if the claim is dismissed, fill in -1.
    
    Just output the formatted judgment list without any comments.
\end{prompt}

\begin{prompt}{Fact-Based LJP}
\small
    \textbf{[Case Type]}

    **one of the ten types**

    \textbf{[Litigant]}

    **the parties concerned**

    \textbf{[Reference Facts]}

    **the fact determined by court**

    \textbf{[Plaintiff's Claims]}

    (1) **claim1**

    (2) **claim2**

    ...

    You need to refer to the evidence presented by all parties in the [Evidence List] and the reference facts provided by the model in the [Reference Facts] to predict the court's judgment on the [Plaintiff's Claims] and form a corresponding judgment list.
    
    The [Judgment List] is a list composed of three numbers (0, 1, -1). If you believe the court will fully support the claim, the result is 1; if partially supported, the result is 0; if the claim is dismissed, fill in -1.
    
    Just output the formatted judgment list without any comments.
\end{prompt}

















    
    

%% file: content/discussion.tex
\section{Discussion}
\label{sec:discuss}


Although the input for LFP is defined as the evidence list and the plaintiff's claims, other trial-related information could also be incorporated into this task. 
As discussed by~\citet{medvedeva2023legal}, the ideal input for LJP should encompass any information available to the court or the parties at the time of performing LJP, such as complaints, defenses, and evidence submitted by the parties. 
This principle also applies to LFP. 
However, the information available to the court or the parties depends on the stage of the trial. 
For example, before filing a lawsuit, the plaintiff and the defendant may only have access to the evidence they personally possess. 
After filing, they gain access to each other's evidence and arguments regarding the legal facts. 
In this work, we chose the evidence list as the basic input for LFP because, at different stages of the trial, both parties have access to certain evidence. 

Note that the "evidence list" here does not necessarily correspond to the final list of evidence submitted to the court, but rather represents the set of evidence items available to the parties at the current stage. 
Additionally, if the parties have access to other trial-related information, it can be incorporated as supplementary input to improve prediction accuracy. 
This suggests that in future work, we can adapt the LFP task to different trial stages by tailoring the input, thereby addressing various demands in legal practice.